\documentclass[11pt]{article}
\usepackage{coling2016}
\usepackage{times}
\usepackage{url}
\usepackage{latexsym}
\usepackage{times}
\usepackage{amsmath}
\usepackage{latexsym}
\usepackage{graphics}
\usepackage{graphicx}
\usepackage{float}
\usepackage{tabularx}
\usepackage{subcaption}
\usepackage[export]{adjustbox}
\DeclareMathOperator*{\argmin}{argmin}

\title{Joint Embedding of Hierarchical Categories and Entities for Concept Categorization and Dataless Classification}
\author{Yuezhang Li, Ronghuo Zheng, Tian Tian, Zhiting Hu, Rahul Iyer, Katia Sycara
\\ Language Technology Institute, School of Computer Science
\\ Carnegie Mellon University
\\{\tt yuezhanl@andrew.cmu.edu}}

\date{}

\begin{document}
\maketitle
\begin{abstract}
Due to the lack of structured knowledge applied in learning distributed representation of categories, existing work cannot incorporate category hierarchies into entity information.~We propose a framework that embeds entities and categories into a semantic space by integrating structured knowledge and taxonomy hierarchy from large knowledge bases. The framework allows to compute meaningful semantic relatedness between entities and categories.~Our framework can handle both single-word concepts and multiple-word concepts with superior performance on concept categorization and yield state of the art results on dataless hierarchical classification.
\end{abstract}

\section{Introduction}

Hierarchies, most commonly represented as Tree or Directed Acyclic Graph (DAG) structures, provide a natural way to categorize and locate knowledge in large knowledge bases (KBs).~For example, WordNet, Freebase and Wikipedia use hierarchical taxonomy to organize entities into category hierarchies. These hierarchical categories could benefit applications such as concept categorization \cite{rothenhausler2009unsupervised}, object categorization \cite{verma2012learning}, document classification \cite{gopal2013recursive}, and link prediction in knowledge graphs \cite{lin2015learning}.~In all of these applications, it is essential to have a good representation of categories and entities as well as a good semantic relatedness measure.

In this paper, we propose two models to simultaneously learn entity and category representations from large-scale knowledge bases (KBs).~They are Category Embedding model and Hierarchical Category Embedding model. The {\bf Category Embedding model} (CE model) extends the entity embedding method of \cite{hu2015entity} by using category information with entities to learn entity and category embeddings.~The {\bf Hierarchical Category Embedding model} (HCE model) extends CE model by integrating categories' hierarchical structure.~It considers all ancestor categories of one entity.~The final learned entity and category vectors can capture meaningful semantic relatedness between entities and categories.

We train the category and entity vectors on Wikipedia, and then evaluate our methods from {\bf two applications: concept\footnote{In this paper, concepts and entities denote same thing.} categorization} \cite{baroni2010distributional} and {\bf dataless hierarchical classification} \cite{song2014dataless}.

\begin{figure*}[h]
    \centering
    \includegraphics[width=\textwidth]{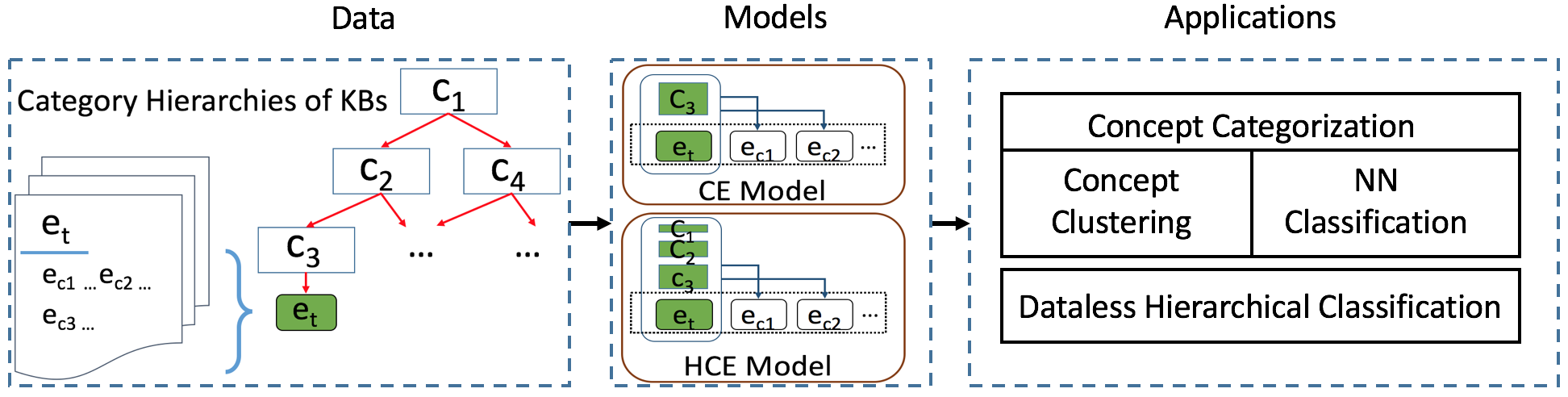}
    \caption{The organization of research elements comprising this paper.}
    \label{fig:1}
\end{figure*}

The organization of the research elements that comprise this paper, summarizing the above discussion, is shown in Figure \ref{fig:1}. 

The main contributions of our paper are summarized as follows. First, we incorporate category information into entity embeddings with the proposed CE model to get entity and category embeddings simultaneously. Second, we add category hierarchies to CE model and develop HCE model to enhance the embeddings' quality. Third, we propose a concept categorization method based on nearest neighbor classification that avoids the issues arising from disparity in the granularity of categories that plague traditional clustering methods.~Fourth, we construct a new concept categorization dataset from Wikipedia. Fifth, we show the potential of utilizing entity embeddings on dataless classification. Overall, our model produce state of the art performance on both concept categorization and dataless hierarchical classification.



\section{Related Work}

\textbf{Relational Embedding}\label{relational embedding}, also known as knowledge embedding, is a family of methods to represent entities as vectors and relations as operations applied to entities such that certain properties are preserved. \cite{paccanaro2001learning,hinton2002learning,nickel2011three,bordes2011learning,nickel2012factorizing,jenatton2012latent,bordes2013translating,neelakantan2015inferring}. For instance, the linear relational embedding \cite{paccanaro2001learning,hinton2002learning} applies a relation to an entity based on matrix-vector multiplication while TransE \cite{bordes2013translating} simplifies the operation to vector addition.~To derive the embedding representation, they minimize a global loss function considering all (entity, relation, entity) triplets so that the embeddings encode meaningful semantics. Our approach is different from this line since we use probabilistic models instead of transition-based models. 

Another line of \textbf{Entity Embedding} methods is based on the skip-gram model \cite{mikolov2013distributed}, a recently proposed embedding model that learns to predict each context word given the target word. This model tries to maximize the average log likelihood of the context word so that the embeddings encode meaningful semantics. For instance, Entity hierarchy embedding \cite{hu2015entity} extends it to predict each context entity given target entity in KBs. \cite{yamada2016joint} proposed a method to jointly embed words and entities through jointly optimizing word-word, entity-word, and entity-entity predicting models. Our models extend this line of research by incorporating hierarchical category information to jointly embed categories and entities in the same semantic space.

\textbf{Category Embedding} has been widely explored with different methods.~\cite{weinberger2009large} initially proposed a taxonomy embedding to achieve document categorization and derive the embeddings of taxonomies of topics in form of topic prototypes.~Furthermore, \cite{hwang2014unified} proposed a discriminative learning framework that can give category embeddings by approximating each category embeddings as a sum of its direct supercategory plus a sparse combination of attributes. However, these methods primarily target documents/object classification not entity representations. 

Recently, several representations are proposed to extend word representation for phrases \cite{yin2014exploration,yu2015learning,passos2014lexicon}. However, they don't use structured knowledge to derive phrase representations.

\section{Joint Embedding of Categories and Entities}
In order to find representations for categories and entities that can capture their semantic relatedness, we use existing hierarchical categories and entities labeled with these categories, and explore two methods: 1) {\bf Category Embedding model} (CE Model): it  replaces the entities in the context with their directly labeled categories to build categories' context; 2) {\bf Hierarchical Category Embedding} (HCE Model): it  further incorporates all ancestor categories of the context entities to utilize the hierarchical information.

\subsection{Category Embedding (CE) Model}\label{CE_model}
Our category embedding (CE) model is based on the Skip-gram word embedding model\cite{mikolov2013distributed}. The skip-gram model aims at generating word representations that are good at predicting \emph{context} words surrounding a \emph{target} word in a sliding window. Previous work \cite{hu2015entity} extends the entity's context to the whole article that describes the entity and acquires a set of entity pairs $\mathbf{D} = \{(e_t, e_c)\}$,  where $e_t$ denotes the {\em target} entity and $e_c$ denotes the {\em context} entity.

Our CE model extends those approaches by incorporating category information.~In KBs such as Wikipedia, category hierarchies are usually given as DAG or tree structures, and entities are categorized into one or more categories as leaves.~Thus, in KBs, each entity $e_t$ is labeled with one or more categories $(c_{1}, c_{2},...,c_{k}), k\geq 1$ and described by an article containing other {\em context} entities (see Data in Figure~\ref{fig:1}).

To learn embeddings of entities and categories simultaneously, we adopt a method that incorporates the labeled categories into the entities when predicting the context entities, similar to TWE-1 model \cite{liu2015topical} which incorporates topic information with words to predict context words. For example, if $e_t$ is the \emph{target} entity in the document, its labeled categories $(c_{1}, c_{2},...,c_{k})$ would be combined with the entity $e_t$ to predict the context entities like $e_{c1}$ and $e_{c2}$ (see CE Model in Figure~\ref{fig:1}).~For each target-context entity pair $(e_t, e_c)$, the probability of $e_c$ being context of $e_t$ is defined as the following softmax:
\begin{equation}
	P(e_c|e_t) = \frac{\exp{(e_t \cdot e_c)}}{\sum_{e \in \mathbf{E}}\exp{(e_t \cdot e)}},
\label{softmax}
\end{equation}
where $\mathbf{E}$ denotes the set of all entity vectors, and $\exp{}$ is the exponential function. For convenience, here we abuse the notation of $e_t$ and $e_c$ to denote a target entity vector and a context entity vector respectively.

Similar to TWE-1 model, We learn the target and context vectors by maximizing the average log probability:
\begin{equation}
	L = \frac{1}{|\mathbf{D}|}\sum_{(e_c,e_t)\in \mathbf{D}} \Big[\log P(e_c|e_t)+\sum_{c_i \in \mathbf{C}(e_t)}\log P(e_c|c_i) \Big],
\end{equation}
where $\mathbf{D}$ is the set of all entity pairs and we abuse the notation of $c_i$ to denote a category vector, and $\mathbf{C}(e_t)$ denotes the categories of entity $e_t$.

\subsection{Hierarchical Category Embedding(HCE) Model}
From the design above, we can get the embeddings of all categories and entities in KBs without capturing the semantics of hierarchical structure of categories. In a category hierarchy, the categories at lower layers will cover fewer but more specific concepts than categories at upper layers. To capture this feature, we extend the CE model to further incorporate the ancestor categories of the target entity when predicting the context entities (see HCE Model in Figure~\ref{fig:1}). If a category is near an entity, its ancestor categories would also be close to that entity. On the other hand,  an increasing distance of the category from the target entity would decrease the power of that category in predicting context entities. Therefore, given the target entity $e_t$ and the context entity $e_c$, the objective is to maximize the following weighted average log probability:
 \begin{equation} \label{eq:3}
 	L =\frac{1}{|\mathbf{D}|}\sum_{(e_c,e_t)\in \mathbf{D}}\Big[\log P(e_c|e_t)+\sum_{c_i \in \mathbf{A}(e_t)}w_i \log P(e_c|c_i)\Big],
 \end{equation}
where $\mathbf{A}(e_t)$ represents the set of ancestor categories of entity $e_t$, and $w_i$ is the weight of each category in predicting the context entity. To implement the intuition that a category is more relevant to its closer ancestor, for example, ``NBC Mystery Movies" is more relevant to ``Mystery Movies" than ``Entertainment",  we set $w_i\propto \frac{1}{l(c_c,c_i)}$ where $l(c_c,c_i)$ denotes the average number of steps going down from category $c_i$ to category $c_c$, and it is constrained with $\sum_i w_i = 1$.

\begin{figure*}[h]
    \centering
    \includegraphics[width = 0.8\textwidth]{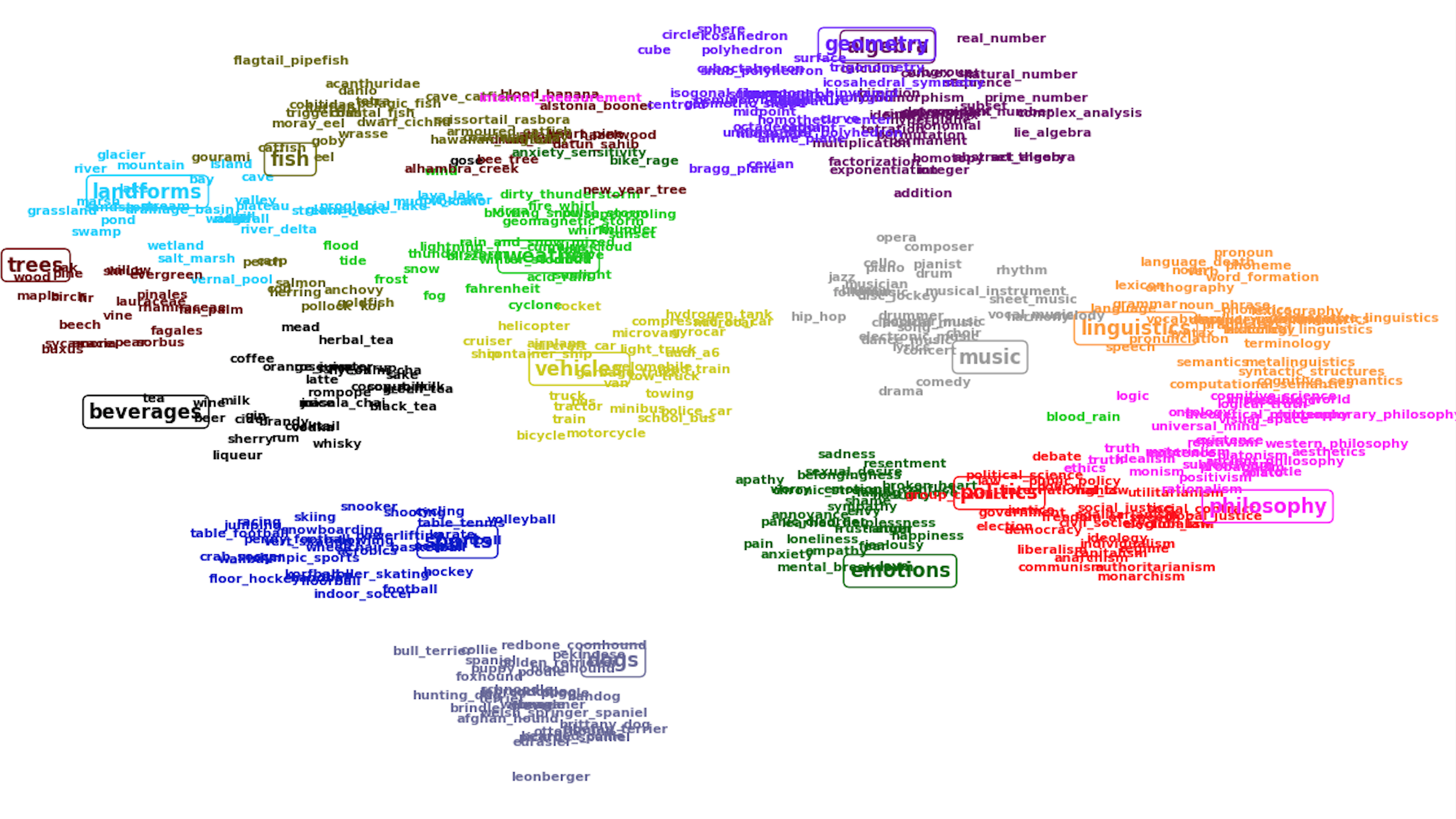}
    \caption{Category and entity embedding visualization of the DOTA-all data set (see Section \ref{ssset:CC_datasets}). We use t-SNE \protect\cite{van2008visualizing} algorithms to map vectors into a 2-dimensional space. Labels with the same color are entities belonging to the same category. Labels surrounded by a box are categories vectors.} 
    \label{vi}
\end{figure*}

Figure \ref{vi} presents the results of our HCE model for DOTA-all data set (see Section \ref{ssset:CC_datasets}). The visualization shows that our embedding method is able to clearly separate entities into distinct categories.

\subsection{Learning}

Learning CE and HCE models follows the optimization scheme of skip-gram model \cite{mikolov2013distributed}. We use negative sampling to reformulate the objective function, which is then optimized through stochastic gradient descent (SGD).

Specifically, the likelihood of each context entity of a target entity is defined with the softmax function in Eq.~\ref{softmax}, which iterates over all entities. Thus, it is computationally intractable. We apply the standard negative sampling technique to transform the objective function in equation (\ref{eq:3}) to equation (\ref{eq:4}) below and then optimize it through SGD:
\begin{align}\label{eq:4}
	L = \sum_{(e_c,e_t)\in \mathbf{D}}\Big[\log\sigma(e_c\cdot e_t) + \sum_{c_i\in \mathbf{A}(e_t)} w_i \log \sigma(e_c \cdot c_i)\Big]+ \\
	\sum_{(e_c', e_t)\in \mathbf{D'}}\Big[\log\sigma(-e_c'\cdot e_t) \notag+\sum_{c_i\in \mathbf{A}(e_t)} w_i \log \sigma(-e_c' \cdot c_i)\Big],
\end{align}
where $\mathbf{D'}$ is the set of negative sample pairs and $\sigma(x)=1/(1+\exp(-x))$ is the sigmoid function.
	
\section{Applications}
We apply our category and entity embedding to two applications: concept categorization and dataless hierarchical classification.

\subsection{Concept Categorization}
Concept\footnote{In this paper, concept and entity denote the same thing.} categorization, also known as \emph{concept learning} or \emph{noun categorization}, is a process of assigning a concept to one candidate category, given a set of concepts and candidate categories.~Traditionally, concept categorization is achieved by {\bf concept clustering} due to the lack of category representations.~Since our model can generate representations of categories, we propose a new method of using {\bf nearest neighbor (NN) classification} to directly categorize each concept to a certain category.

The \textbf{concept clustering} is defined as: given a set of concepts like \emph{dog, cat, apple} and the corresponding gold standard categorizations like \emph{animal, fruit}, apply a word space model to project all the concepts to a semantic space and perform clustering. The clustering results can be evaluated by comparing with the gold standard categorizations. Since we have representation of categories, we propose {\bf nearest neighbor (NN) classification} to categorize concepts by directly comparing concept vectors with candidate category vectors.~Precisely, given a set of concepts $\mathbf{E}$ and a set of candidate categories $\mathbf{C}$, we convert all concepts to concept vectors and all candidate categories to category vectors.~Then we use the equation $c = \argmin_{c_i \in \mathbf{C}}||c_i - e||$ to assign the concept vector $e$ with category $c$. 

Since purity works as a standard evaluation metric for clustering \cite{rothenhausler2009unsupervised}, to compare our model with the concept clustering, we also use purity to measure our model's performance.~Specifically, \textbf{purity} is defined as:
\begin{equation} \label{eq:purity}
\text{purity}(\mathbf{\Omega},\mathbf{G})=\frac{1}{n}\sum_{k}{\max_{j}{|\omega_{k} \cap g_j}|},
\end{equation}
where $\mathbf{\Omega}$ denotes a clustering solution of $n$ clusters, $\mathbf{G}$ is a set of gold standard classes, $k$ is the cluster index and $j$ is the class index, $\omega_k$ represents the set of labels in a cluster and $g_j$ is the set of labels in a class. ~A higher purity indicates better model performance.

\subsection{Dataless Classification}
Dataless classification uses the similarity between documents and labels in an enriched “semantic” space to determine in which category the given document is. It has been proved that explicit semantic analysis (ESA) \cite{gabrilovich2007computing} has shown superior performance on dataless classification \cite{song2014dataless}. ESA uses a bag-of-entities retrieved from Wikipedia to represent the text. For example, the document \textit{"Jordan plays basketball"} can be represented as a ESA sparse vector of
\textit{\{Michael Jordan:48.3, Basketball:29.8, Air Jordan: 28.8, Outline of basketball: 28.5, Chicago Bulls: 23.6\}} in an ESA implementation. After converting documents and short label descriptions to ESA vectors, a nearest neighbor classifier is applied to assign labels for each document.

Due to sparsity problem of ESA vectors, sparse vector densification is introduced to augment the similarity calculation between two ESA vectors \cite{song2015unsupervised}. This is achieved by considering pairwise similarity between entities in ESA vectors. However, they simply use word2vec \cite{mikolov2013distributed} to derive entity representation. We extend it by directly applying entity embeddings and show the potential of entity embeddings to improve dataless classification.

We use averaged $F_1$ scores to measure the performance of all methods \cite{yang1999evaluation}. Let $TP_i$, $FP_i$, $FN_i$ denote the true-positive, false-positive and false-negative values for the $i$th label in label set $T$, the micro- and macro-averaged $F_1$ scores are defined as: $MicroF_1 = 2\overline{P} * \overline{R} /(\overline{P} + \overline{R})$ and $MacroF_1=\frac{1}{|T|}\sum_i^{|T|}\frac{2P_i*R_i}{P_i+R_i}$, where $P_i=TP_i/(TP_i+FP_i)$ and $R_i=TP_i/(TP_i+FN_i)$ are precision and recall for $i$th label, $\overline{P}=\sum_i^{|T|}TP_i/\sum_i^{|T|}(TP_i+FP_i)$ and $\overline{R}=\sum_i^{|T|}TP_i/\sum_i^{|T|}(TP_i+FN_i)$ are averaged precision and recall for all labels.

\section{Experiments}
In the experiments, we use the dataset collected from Wikipedia on Dec.~1, 2015\footnote{https://dumps.wikimedia.org/wikidatawiki/20151201/} as the training data. We preprocess the category hierarchy by pruning administrative categories and deleting bottom-up edges to construct a DAG. The final version of data contains 5,373,165 entities and 793,856 categories organized as a DAG with a maximum depth of 18. The root category is ``main topic classifications".~We train category and entity vectors in dimensions of {50, 100, 200, 250, 300, 400, 500}, with batch size $B=500$ and negative sample size $k=10$.

With the training dataset defined above, we conduct experiments on two applications:~concept categorization and dataless hierarchical classification.

\subsection{Concept Categorization}

\subsubsection{Datasets}\label{ssset:CC_datasets}
There are two datasets used in this experiment. The first one is the {\bf Battig} test set introduced by \cite{baroni2010distributional}, which includes 83 concepts from 10 categories. The {\bf Battig} test set only contains single-word concepts without any multiple-word concepts (e.g., ``table tennis"). Hence, using this dataset restricts the power of concept categorization to single-word level. We use this dataset because it has been used as a benchmark for most previous approaches for concept categorization.

Due to the limitations of the {\bf Battig} test set, we construct a new entity categorization dataset {\bf DOTA }(Dataset Of enTity cAtegorization) with 450 entities categorized into 15 categories (refer to Appendix A).~All the categories and entities are extracted from Wikipedia, so the resulting dataset does not necessarily contains only single-word entities.~Thus, the dataset can be split into two parts, {\bf DOTA-single} that contains 300 single-word entities categorized into 15 categories and {\bf DOTA-mult} that contains 150 multiple-word entities categorized into the same 15 categories.~We design the {\bf DOTA} dataset based on the following principles:

\textbf{Coverage vs Granularity}: Firstly, the dataset should cover at least one category of Wikipedia's main topics including ``Culture", ``Geography", ``Health", ``Mathematics", ``Nature", ``People", ``Philosophy", ``Religion", ``Society" and ``Technology".~Secondly, categories should be in different granularity, from large categories (e.g.,``philosophy") to small categories (e.g., ``dogs").~Large categories are ones that are located within 5 layers away from the root, medium categories are 6-10 layers away from the root, while small categories have distance of 11-18 to the root.~Our dataset consists of 1/3 large categories, 1/3 medium categories, and 1/3 small categories.

\textbf{Single-Words vs Multiple-Words}: Previous concept categorization datasets only contain single-words.~However, some concepts are multiple-words and cannot be simply represented by single-words.~For example, the concept ``hot dog" is very different from the concept ``dog".~Therefore, we make each category of the dataset contain 10 multiple-word entities and 20 single-word entities.

\subsubsection{Baselines}\label{subsubsec:DC_baselines}
We compare our entity and category embeddings with the following state of the art word and entity embeddings.

{\bf WE$_{Mikolov}$}\cite{mikolov2013distributed}: We trained word embeddings with Mikolov's word2vec toolkit\footnote{https://code.google.com/archive/p/word2vec/} on the same Wikipedia corpus as ours (1.7 billion tokens) and then applied the Skip-gram model with negative sample size of 10 and window size of 5 in dimensionality of {50, 100, 200, 250, 300, 400, 500}. 

{\bf WE$_{Senna}$} \cite{collobert2011natural}: We downloaded this 50-dimension word embedding\footnote{http://ronan.collobert.com/senna/} trained on Wikipedia over 2 months.~We use this embedding as a baseline because it is also trained on Wikipedia.

Given the above semantic representation of words, we derive each entity/category embedding by averaging all word vectors among each entity/category label. If the label is a phrase that has been embedded (e.g., Mikolov's embedding contains some common phrases), we direct use the phrase embedding.

{\bf HEE} \cite{hu2015entity}: This Hierarchical Entity Embedding method uses the whole Wikipedia hierarchy to train entity embeddings and distance metrics. We used the tools provided by the authors to train entity embeddings on the same Wikipedia corpus as ours. We set the batch size B = 500, the initial learning rate $\eta$ = 0.1 and decrease it by a factor of 5 whenever the objective value does not increase, and the negative sample size k = 5 as suggested by the authors. This method doesn't provide a way of learning category embeddings. Therefore, it cannot be used in the \textbf{nearest neighbor classification} method.
	
{\bf TransE} \cite{bordes2013translating}: It is a state of the art relational embedding method introduced in Section.~\ref{relational embedding}, which embeds entities and relations at the same time. It can be extended to derive category embeddings, since categories can be seen as a special entity type. To make comparisons fair enough, we adopt three different versions of TransE to derive entity and category embedding to compare with our methods. \textbf{TransE$_1$}: Use entities and categories as two entity types and the direct-labeled-category relation. So the triplets we have is in the form of (entity, direct-labeled-category, category). \textbf{TransE$_2$}: Add to TransE$_1$ with the hierarchy structure of wikipedia categories. Namely, add the super-category relation between categories. Therefore we have new triplets in the form of (category, super-category, category). \textbf{TransE$_3$}: To make it use full information as ours, we extend TransE$_2$ to utilize context entity relationship described in Section. \ref{CE_model}. Therefore we add new triplets in the form of (entity, context-entity, entity). 

\textbf{HC}: To further evaluate the advantage of utilizing category hierarchy in training entity and category embedding, we also compare our Hierarchical Category Embedding (HCE) model with our Category Embedding (CE) model that has no hierarchical information. 

\subsubsection{Results}\label{subsubsec:CC_result}

In the experiments, we used scikit-learn \cite{pedregosa2011scikit} to perform clustering. We tested k-means and hierarchical clustering with different distance metrics (euclidean, cosine) and linkage criterion (ward, complete, average). All these choices along with the vector dimensionality are treated as our models' hyper-parameters. For selecting hyper-parameters, we randomly split the Battig and Dota datasets to 50\% of validation data and 50\% of test data, evenly across all categories. We trained all the embeddings (except SENNA) on the same Wikipedia dump and tuned hyper-parameters on the validation set. For experiments on Dota dataset, since the ground truth is contained in our Wikipedia corpus, we deleted all category-entity links contained in Dota dataset from our category hierarchy to train HEE, TransE and HCE embeddings to make comprison fair enough.

Table.~\ref{purity1} shows the experimental results of the {\bf concept clustering} method.~It is clear that {\bf hierarchical category embedding} (HCE) model outperforms other methods in all datasets.~Our model achieves a purity of 89\% on Battig and 89\% on DOTA-all.

\begin{table}[h]
\centering
\begin{subtable}{.45\linewidth}
\resizebox{\columnwidth}{!}{
\begin{tabular}{|c|c|c|c|c|}
\hline
               & \textbf{Battig} & \textbf{DOTA-single} & \textbf{DOTA-mult} & \textbf{DOTA-all} \\ \hline
WE$_{Senna}$ (50,50)   & 0.74           & 0.61                & 0.43              & 0.45             \\ \hline
WE$_{Mikolov}$ (400,400) & 0.86           & 0.83                & 0.73              & 0.78             \\ \hline
HEE (400,400)           & 0.82           & 0.83                & 0.80              & 0.81             \\ \hline
TransE$_1$ (300,250)     & 0.67           & 0.71                & 0.68              & 0.69             \\ \hline
TransE$_2$ (400,300)    & 0.73           & 0.78                & 0.75           & 0.76             \\ \hline
TransE$_3$ (200,400)    & 0.43           & 0.53                & 0.50            & 0.51             \\ \hline
CE (300,200)            & 0.84           & 0.86                & 0.83              & 0.85             \\ \hline
HCE (400,400)       & \textbf{0.89}   & \textbf{0.92}       & \textbf{0.88}      & \textbf{0.89}     \\ \hline
\end{tabular}
}
\caption{Purity of {\bf concept clustering} method}
\label{purity1}
\end{subtable}
\hspace{2em}
\begin{subtable}{.45\linewidth}
\resizebox{\columnwidth}{!}{
\begin{tabular}{|c|c|c|c|c|}
\hline
               & \textbf{Battig} & \textbf{DOTA-single} & \textbf{DOTA-mult} & \textbf{DOTA-all} \\ \hline
WE$_{Senna}$ (50,50)   & 0.44           & 0.52                & 0.32              & 0.45             \\ \hline
WE$_{Mikolov}$ (400,400) & 0.74           & 0.74                & 0.67              & 0.72             \\ \hline
HEE           & -           & -                & -              & -             \\ \hline
TransE$_1$ (300,250)     & 0.66           & 0.72                & 0.69              & 0.71             \\ \hline
TransE$_2$ (400,300)    & 0.75           & 0.80                & 0.77          & 0.79             \\ \hline
TransE$_3$ (300,400)     & 0.46           & 0.55                & 0.52              & 0.54             \\ \hline
CE (200,200)         & 0.79           & 0.89                & 0.85              & 0.88             \\ \hline
HCE (400,400)         & \textbf{0.87}   & \textbf{0.93}        & \textbf{0.91}      & \textbf{0.92}     \\ \hline
\end{tabular}
}
\caption{Purity of {\bf NN classification} method}
\label{purity2}
\end{subtable}
\caption{Purity of {\bf nearest neighbor (NN) classification} and  {\bf concept clustering} methods with different embeddings on Battig and DOTA datasets. The two numbers given in each parentheses are the vector dimensionality among \{50,100,200,250,300,400,500\} that produce the best results on Battig and Dota validation set respectively. All the purity scores are calculated under such choices of hyper-parameters on the test set.}
\end{table}

For word embeddings like Mikolov and Senna, the performance drops a lot from single-word entity categorization to multiple-word entity categorization, because these embeddings mainly contain single words.~To get the embeddings of multiple-word, we use the mean word vectors to denote multiple-word embeddings.~ However, the meaning of a multiple-word is not simply the aggregation of the meaning of the words it contains. The good performance of HEE shows the high quality of its entity embeddings. By incorporating category hierarchy information, TransE$_2$ gets an advantage over TransE$_1$. This advantage can also be observed when we incorporate category hierarchy information into CE. However, further incorporating of context entities deteriorates the performance of TransE. This may due to the nature of transition-based models that assume relationships as transitions among entities. They get much noise when introduced with many context entity triplets along with only one entity-to-context-entity relation type.

Table.\ref{purity2} shows the experimental results of the {\bf nearest neighbor (NN) classification} method. The results indicate the feasibility of using category vectors to directly predict the concept categories without clustering entities. Our model achieves a purity of 87\% on Battig and 92\% on DOTA-all.

\subsection{Dataless Classification}
\subsubsection{Datasets}
{\bf 20Newsgroups Data(20NG):}The 20 newsgroups dataset \cite{lang1995newsweeder} contains about 20,000 newsgroups documents evenly categorized to 20 newsgroups, and further categorized to six super-classes. We use the same label description provided by \cite{song2014dataless}.

{\bf RCV1 Dataset:} The RCV1 dataset \cite{lewis2004rcv1} contains 804,414 manually labeled newswire documents, and categorized with respect to three controlled vocabularies: industries, topics and regions. We use topics as our hierarchical classification problem. There are 103 categories including all nodes except for the root in the hierarchy, and the maximum depth is 4. To ease the computational cost of comparison, we follow the chronological split proposed in \cite{lewis2004rcv1} to use the first 23,149 documents marked as training samples in the dataset. The dataset also provides the name and the description of each category label.

\subsubsection{Implementation Details}
The baseline of similarity measure between each label and each document is cosine similarity between corresponding ESA vectors. We use ESA with 500 entities, and augment similarity measure by plugging in different embeddings introduced in Section.~\ref{subsubsec:DC_baselines} with Hungarian sparse vector densification method described in \cite{song2015unsupervised}.  The Hungarian method is a combinatorial optimization algorithm aiming to find an optimal assignment matching the two sides of a bipartite graph on a one-to-one basis. We average over all cosine similarities of entity pairs produced by the Hungarian method to produce similarity between ESA vectors. In addition, we cut off all entity similarities below 0.85 and map them to zero empirically.

For classification, We use the bottom-up pure dataless hierarchical classification algorithm which proved to be superior in \cite{song2014dataless} and set the threshold $\delta$ to be 0.95 empirically.
\subsubsection{Results}
\begin{figure}[h]
    \centering
    \begin{subfigure}[]{0.48\textwidth}
        \centering
        \includegraphics[width=\textwidth]{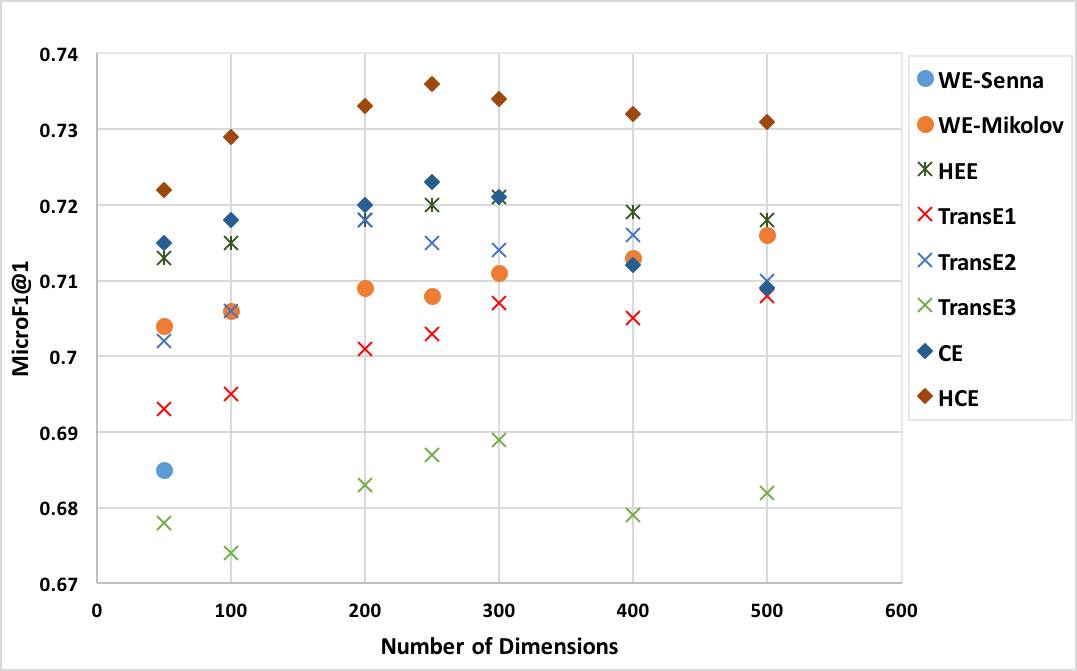}
        \caption{20newsgroups: bottom-up (ESA:0.682)}
        \label{fig:20news}
    \end{subfigure}~~
    \begin{subfigure}[]{0.48\textwidth}
        \centering
        \includegraphics[width=\textwidth]{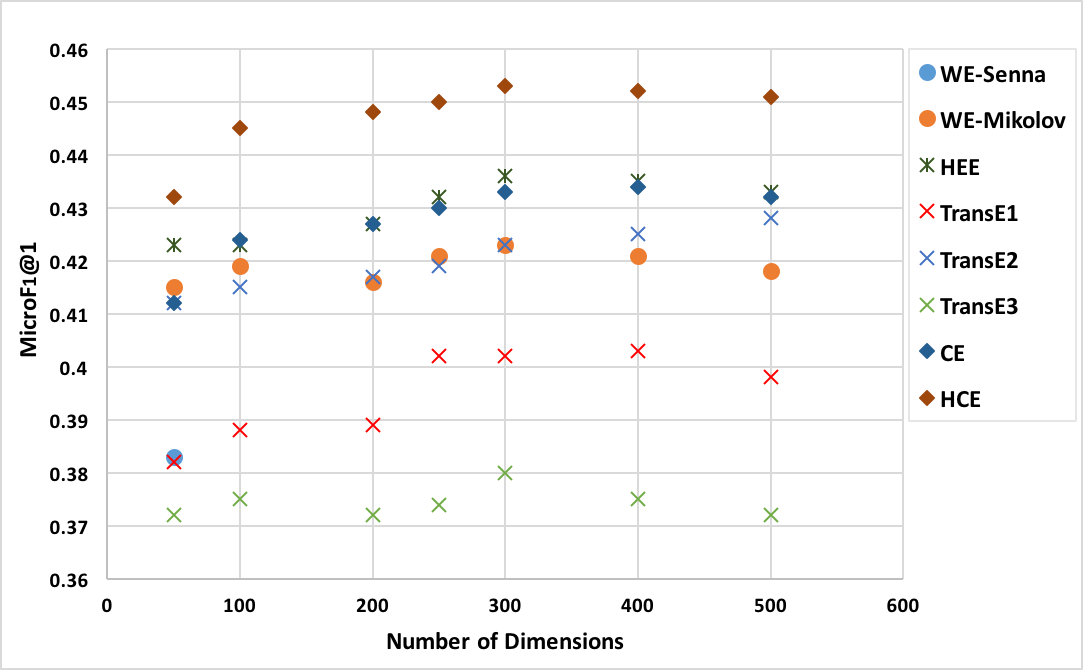}
        \caption{RCV1: bottom-up (ESA:0.371)}
    \end{subfigure}
    
    \caption{MicroF$_1$@1 results of ESA augmented with different embeddings for dataless hierarchical classification. It is clear the ESA densified with HCE performs best on dataless hierarchical classification.}
    \label{dl}
\end{figure}
We perform dataless hierarchical classification using the same ESA settings, densified with different embedding methods for similarity calculation. Figure.~\ref{dl} presents the classification performance with different embedding methods in dimensionality of \{50,100,200,250,300,400,500\}. It is clear that ESA densified with HCE stably outperform other competitive methods on these two datasets, which indicates the high quality of entity embeddings derived by HCE.

Besides, we observe that entity embedding methods such as TransE$_2$, HEE and HCE perform better than word embedding methods. This is because ESA represents text as bag-of-entities, and entity embeddings successfully encode entity similarities. Compared with bag-of-words baselines in \cite{song2014dataless}, it is clear that bag-of-entities models can better capture semantic relatedness between documents and short category descriptions.

\section{Conclusion}
In this paper, we proposed a framework to learn entity and category embeddings to capture semantic relatedness between entities and categories. This framework can incorporate taxonomy hierarchy from large scale knowledge bases. Experiments on both concept categorization and dataless hierarchical classification indicate the potential usage of category and entity embeddings on more other NLP applications. 

\bibliography{coling2016}
\bibliographystyle{acl2016}

\newpage
\onecolumn
\appendix
\section{The DOTA dataset: 300 single-word entities and 150 multi-word entities from 15 Wikipedia Categories}
\begin{table*}[h]
\small
\centering
\label{my-label}
\begin{tabular}{|l|p{14cm}|}
\hline
\textbf{Category}   & \textbf{Entities}                                                                                                                                                                                                                                                                                                                                                                                              \\ \hline
beverages  & juice, beer, milk, coffee, tea, cocktail, wine, liqueur, sake, vodka, mead, sherry, brandy, gin, rum, latte, whisky, cider, gose, rompope, orange juice, masala chai, green tea, black tea, herbal tea, coconut milk, corn syrup, soy milk, rose water, hyeonmi cha                                                                                                                                                             \\ \hline
sports     & bowling, football, aerobics, hockey, karate, korfball, handball, floorball, skiing, cycling, racing, softball, shooting, netball, snooker, powerlifting, jumping, wallball, volleyball, snowboarding, table tennis, floor hockey, olympic sports, wheelchair basketball, crab soccer, indoor soccer, table football, roller skating, vert skating, penny football                                                                  \\ \hline
emotions   & love, anxiety, empathy, fear, envy, loneliness, shame, anger, annoyance, happiness, jealousy, apathy, resentment, frustration, belongingness, sympathy, pain, worry, hostility, sadness, broken heart, panic disorder, sexual desire, falling in love, emotional conflict, learned helplessness, chronic stress, anxiety sensitivity, mental breakdown, bike rage                                                                  \\ \hline
weather    & cloud, wind, thunderstorm, fog, snow, wave, blizzard, sunlight, tide, virga, lightning, cyclone, whirlwind, sunset, dust, frost, flood, thunder, supercooling, fahrenheit, acid rain, rain and snow mixed, cumulus cloud, winter storm, blowing snow, geomagnetic storm, blood rain, fire whirl, pulse storm, dirty thunderstorm                                                                                                  \\ \hline
landforms  & lake, waterfall, stream, river, wetland, marsh, valley, pond, sandstone, mountain, cave, swamp, ridge, plateau, cliff, grassland, glacier, hill, bay, island, glacial lake, drainage basin, river delta, stream bed, vernal pool, salt marsh, proglacial lake, mud volcano, pit crater, lava lake                                                                                                                                  \\ \hline
trees      & wood, oak, pine, evergreen, willow, vine, shrub, birch, beech, maple, pear, fir, pinales, lauraceae, sorbus, buxus, acacia, rhamnaceae, fagales, sycamore, alhambra creek, alstonia boonei, atlantic hazelwood, bee tree, blood banana, datun sahib, druid oak, new year tree, heart pine, fan palm                                                                                                                                \\ \hline
algebra    & addition, multiplication, exponentiation, tetration, polynomial, calculus, permutation, subgroup, integer, monomial, bijection, homomorphism, determinant, sequence, permanent, homotopy, subset, factorization, associativity, commutativity, real number, abstract algebra, convex set, prime number, complex analysis, natural number, complex number, lie algebra, identity matrix, set theory
\\ \hline
geometry   & trigonometry, circle, square, polyhedron, surface, sphere, cube, icosahedron, hemipolyhedron, digon, midpoint, centroid, octadecagon, curvature, curve, zonohedron, cevian, orthant, cuboctahedron, midsphere, regular polygon, uniform star polyhedron, isogonal figure, icosahedral symmetry, hexagonal bipyramid, snub polyhedron, homothetic center, geometric shape, bragg plane, affine plane
                               \\ \hline
fish       & goldfish, gourami, koi, cobitidae, tetra, goby, danio, wrasse, acanthuridae, anchovy, carp, catfish, cod, eel, flatfish, perch, pollock, salmon, triggerfish, herring, cave catfish, coachwhip ray, dwarf cichlid, moray eel, coastal fish, scissortail rasbora, flagtail pipefish, armoured catfish, hawaiian flagtail, pelagic fish                                                                                            \\ \hline
dogs       & spaniel, foxhound, bloodhound, beagle, pekingese, weimaraner, collie, terrier, poodle, puppy, otterhound, labradoodle, puggle, eurasier, drever, brindle, schnoodle, bandog, leonberger, cockapoo, golden retriever, tibetan terrier, bull terrier, welsh springer spaniel, hunting dog, bearded collie, picardy spaniel, afghan hound, brittany dog, redbone coonhound                                                           \\ \hline
music      & jazz, blues, song, choir, opera, rhythm, lyrics, melody, harmony, concert, comedy, violin, drum, piano, drama, cello, composer, musician, drummer, pianist, hip hop, classical music, electronic music, folk music, dance music, musical instrument, disc jockey, popular music, sheet music, vocal music                                                                                                                          \\ \hline
politics   & democracy, law, government, liberalism, justice, policy, rights, utilitarianism, election, capitalism, ideology, egalitarianism, debate, regime, globalism, authoritarianism, monarchism, anarchism, communism, individualism, freedom of speech, political science, public policy, civil society, international law, social contract, election law, social justice, global justice, group conflict                                \\ \hline
philosophy & ethics, logic, ontology, aristotle, plato, rationalism, platonism, relativism, existence, truth, positivism, metalogic, subjectivism, idealism, materialism, aesthetics, probabilism, monism, truth, existence, western philosophy, contemporary philosophy, cognitive science, logical truth, ancient philosophy, universal mind, visual space, impossible world, theoretical philosophy, internal measurement
                    \\ \hline
linguistics & syntax, grammar, semantics, lexicon, speech, phonetics, vocabulary, phoneme, lexicography, language, pragmatics, orthography, terminology, pronoun, noun, verb, pronunciation, lexicology, metalinguistics, paleolinguistics, language death, historical linguistics, dependency grammar, noun phrase, comparative linguistics, word formation, cognitive semantics, syntactic structures, auxiliary verb, computational semantics \\ \hline
vehicles   & truck, car, aircraft, minibus, motorcycle, microvan, bicycle, tractor, microcar, van, ship, helicopter, airplane, towing, velomobile, rocket, train, bus, gyrocar, cruiser, container ship, school bus, road train, tow truck, audi a6, garbage truck, hydrogen tank, light truck, compressed air car, police car                                                                                                                  \\ \hline

\end{tabular}
\end{table*}

\end{document}